\documentclass{article}
\usepackage[margin=1in]{geometry}
\usepackage{amsmath,amsfonts}
\usepackage{graphicx}
\usepackage{hyperref}
\usepackage{booktabs}
\usepackage{natbib}
\usepackage{times}

\title{Temporal Entailment Pretraining for Clinical Language Models over EHR Data}
\author{Tatsunori Tanaka, Fi Zheng, Kai Sato, Zhifeng Li, Yuanyun Zhang, Shi Li}

\begin{document}
\maketitle

\begin{abstract}
Clinical language models have achieved strong performance on downstream tasks by pretraining on domain-specific corpora such as discharge summaries and medical notes. However, most approaches treat the electronic health record (EHR) as a static document, neglecting the temporally-evolving and causally-entwined nature of patient trajectories. In this paper, we introduce a novel temporal entailment pretraining objective for language models in the clinical domain. Our method formulates EHR segments as temporally ordered sentence pairs and trains the model to determine whether a later state is entailed by, contradictory to, or neutral with respect to an earlier state. Through this temporally-structured pretraining task, models learn to perform latent clinical reasoning over time, improving their ability to generalize across forecasting and diagnosis tasks. We pretrain on a large corpus derived from MIMIC-IV and demonstrate state-of-the-art results on temporal clinical QA, early warning prediction, and disease progression modeling.
\end{abstract}

\section{Introduction}

Electronic health records (EHRs) \citep{seymour2012electronic, kim2019evolving} are among the most temporally and semantically dense data sources available for machine learning \citep{rajpurkar2022ai}. A single patient's trajectory may span years of clinical encounters, capturing symptoms, diagnostics, interventions, and outcomes, each timestamped and variably documented in structured and unstructured formats \citep{zhao2019learning, huguet2020using}. This makes EHRs a fertile ground for developing temporally-aware machine learning models that can reason about patient states, anticipate outcomes, and provide assistive decision support. Yet, despite this richness, current language models trained on EHR data often fail to incorporate the intrinsic temporality of the clinical domain in a meaningful way.

The prevailing paradigm in clinical NLP has relied heavily on masked language modeling (MLM) objectives \citep{salazar2019masked}, adapted from general-purpose models like BERT \citep{devlin2019bertpretrainingdeepbidirectional}. While effective for learning surface-level co-occurrence statistics, MLM does not naturally encode directional or causal relationships across time, nor does it encourage the model to understand how earlier clinical states influence later outcomes. As a result, downstream tasks that require longitudinal reasoning—such as forecasting disease progression, generating temporal summaries, or performing early warning detection—often require task-specific architectures or external supervision.

In contrast to these approaches, we argue that clinical language models should be explicitly trained to reason over time, not merely across context windows. Just as human clinicians form diagnostic impressions based on how symptoms evolve or resolve over time, so too should models be equipped to interpret EHRs as temporally ordered sequences where earlier events imply, contradict, or are agnostic to subsequent ones. This motivates a shift away from static snapshot modeling and toward representations that encode temporal entailment.

Drawing inspiration from natural language inference (NLI)—a central task in general-domain NLP that captures semantic relationships between text pairs—we introduce a new self-supervised objective tailored to the clinical domain: \textbf{Temporal Entailment Pretraining (TEP)}. Rather than predicting masked tokens or next notes, we construct sentence pairs from temporally-separated segments of the same patient's record. The model must determine whether the later segment is logically entailed by, contradicts, or is neutral with respect to the earlier segment. Unlike standard NLI, which depends on curated corpora such as SNLI or MNLI \citep{bowman2015large, williams2018broad}, our supervision is derived automatically from clinical temporal patterns, leveraging weak heuristics, diagnostic staging progressions, and synthetic augmentation.

This formulation yields several advantages. First, it imposes a directional, asymmetric relationship between input pairs, mirroring the causal and diagnostic dependencies clinicians often infer. Second, it induces temporally-aligned representation spaces that naturally capture clinical trajectories \citep{steinberg2023motortimetoeventfoundationmodel}. Third, it generalizes across data modalities: the EHR segments can encode free-text notes \citep{huang2019empirical, gabriel2018identifying}, structured codes \citep{lee2024can}, lab-derived summaries \citep{hegselmann2023tabllmfewshotclassificationtabular, lee2024emergency, ono2024text}, or even mixed representations (multimodal) \citep{yang2021leverage}. Finally, it scales, enabling pretraining over millions of temporal pairs without manual annotation.

Through TEP, we bridge the gap between temporal reasoning and language model pretraining in the clinical domain. We demonstrate that models trained with this objective not only improve performance on traditional clinical NLP benchmarks but also exhibit stronger generalization and temporal calibration on tasks requiring sequential prediction. Ultimately, our work suggests a new direction for language model development in healthcare: one where temporality is a core part of the pretraining signal, not an afterthought.

\section{Related Work}

\subsection{Clinical Language Models}

The development of domain-specific language models has significantly advanced clinical natural language processing (NLP). Early efforts, such as ClinicalBERT \citep{alsentzer2019publicly}, adapted BERT to the clinical domain by fine-tuning on MIMIC-III notes \citep{johnson2016mimic}, resulting in improved performance on tasks like named entity recognition and medical natural language inference (MedNLI). Building upon this, Clinical ModernBERT \citep{lee2025clinical} introduced architectural enhancements from modernbert \citep{warner2024smarterbetterfasterlonger}, including rotary positional embeddings and Flash Attention, and extended context lengths up to 8,192 tokens. Pretrained on a diverse corpus comprising PubMed abstracts, MIMIC-IV clinical data, and medical ontologies, Clinical ModernBERT demonstrated superior performance on long-context biomedical NLP tasks, reaffirming the efficacy of encoder-only architectures in clinical settings. There are other variants with other modifications, some of which we will highlight here: MedBERT \citep{liu2021med}, ClinicalBERT \citep{huang2019clinicalbert}, Clincial Longformer \citep{li2022clinicallongformerclinicalbigbirdtransformerslong}.

\subsection{Natural Language Inference in Clinical Contexts}

Natural Language Inference (NLI) has emerged as a critical task in clinical NLP, enabling models to reason about entailment relationships between clinical statements \citep{herlihy-rudinger-2021-mednli}. The MedNLI dataset \citep{romanov-shivade-2018-lessons} was specifically designed to evaluate NLI in the clinical domain, highlighting the challenges posed by domain-specific language and the necessity for specialized models. Subsequent studies have leveraged transformer-based architectures, such as ClinicalBERT \citep{huang2019clinicalbert}, to improve NLI performance on MedNLI, underscoring the importance of domain adaptation in clinical reasoning tasks.

\subsection{Temporal Modeling in Clinical NLP}

Temporal dynamics are intrinsic to clinical narratives, yet many models inadequately capture this aspect. Recent approaches have sought to integrate temporal reasoning into clinical NLP models. For instance, methods that model disease progression or predict future clinical events necessitate an understanding of temporal relationships within patient records. 

We will acknowledge that aside from Clinical NLP methods, there has been a lot of work in the EHR foundation model space \citep{wornow2023shaky} that has effectively integrated temporal aspects into the modeling framework \citep{mcdermott2023event, steinberg2023motortimetoeventfoundationmodel, pang2021cehr}. However in the natural language format, this is much more difficult to represent. Our proposed Temporal Entailment Pretraining (TEP) framework builds upon these efforts by explicitly modeling temporal entailment between patient states, thereby enhancing the model's capacity for longitudinal clinical reasoning.

\section{Method}

\subsection{Problem Setting and Temporal Semantics}

Let $\mathcal{P}$ denote the set of patients and for each patient $p \in \mathcal{P}$, let $\mathcal{E}_p = \{(x_t, t) \}_{t \in T_p}$ be a temporally ordered sequence of clinical events, where $x_t$ is a textual representation of the patient's state at time $t$. This representation may consist of clinical notes, structured diagnosis and procedure codes, medication records, lab summaries, or generated natural language descriptions. The goal is to model entailment relations between temporally separated clinical observations, such that given a pair $(x_t, x_{t'})$ with $t < t'$, the model predicts whether $x_{t'}$ is entailed by $x_t$, contradicts it, or is temporally independent (neutral).

\subsection{Temporal Entailment as Clinical Reasoning}

We posit that clinical reasoning can be framed as temporal inference: a model must determine whether the future (e.g., a later diagnosis, lab result, or intervention) logically follows from a known clinical past. More formally, given $(x_t, x_{t'})$, we define an entailment relation:
\[
f(x_t, x_{t'}) = y \in \mathcal{Y}, \quad \text{where} \quad \mathcal{Y} = \{\texttt{entail}, \texttt{contradict}, \texttt{neutral}\}
\]
This framing aligns closely with the inferential process undertaken by clinicians, who synthesize patient history to anticipate or rule out future states. Importantly, the directionality ($t < t'$) imposes an asymmetry that distinguishes our task from static textual entailment, requiring temporal causal inference.

\subsection{Data Construction and Weak Labeling}

To construct training pairs $(x_t, x_{t'})$, we segment patient timelines into temporally bounded windows of clinical activity. Sentences or aggregated embeddings are derived from structured and unstructured sources using templates and representation functions $\phi: \text{raw EHR} \rightarrow \mathcal{X}$. Weak supervision \citep{dehghani2017neural} is applied to assign $y$ using heuristics derived from diagnostic ontologies \citep{steinberg2023motortimetoeventfoundationmodel}, UMLS progression patterns, and known pathophysiological sequences. For example, progression from "stage 2 chronic kidney disease" to "stage 4 CKD" entails a worsening trajectory (entail), while regression to "normal GFR" contradicts the earlier state (contradict). Transitions between orthogonal systems (e.g., dermatological and neurological) are labeled as neutral.

\subsection{Model Architecture}

We adopt a transformer-based encoder $f_{\theta}: \mathcal{X} \rightarrow \mathbb{R}^d$ to compute contextual embeddings for each segment. To inject temporal inductive bias, we replace absolute position embeddings with Rotary Positional Embeddings (RoPE) \citep{su2023roformerenhancedtransformerrotary}, which encode relative and cyclic dependencies:
\[
\text{RoPE}(x_t) = x_t \odot R(t), \quad \text{where } R(t) \in \mathbb{C}^{d} \text{ is a rotation matrix parameterized by time}
\]
This allows fine-grained encoding of temporal offsets between $x_t$ and $x_{t'}$ within a fixed-length input, preserving alignment across windows.

The two inputs $(x_t, x_{t'})$ are jointly encoded via cross-attention in a bi-sequence setup:
\[
h = \text{Transformer}([\texttt{CLS}, x_t, \texttt{SEP}, x_{t'}])
\]
where $h_{\texttt{CLS}}$ is passed through a classification head $\psi: \mathbb{R}^d \rightarrow \mathbb{R}^{|\mathcal{Y}|}$ to predict the entailment relation.

\subsection{Contrastive Objective}

We define a multi-class classification loss over the entailment label:
\[
\mathcal{L}_{\text{TEP}}(\theta) = - \sum_{(x_t, x_{t'}, y) \in \mathcal{D}} \log \text{softmax}(\psi(f_\theta(x_t, x_{t'})))_y
\]
In practice, we incorporate label smoothing to prevent overfitting to noisy weak labels and regularize predictions across temporally adjacent triplets.

\subsection{Theoretical Motivation and Representation Geometry}

The proposed TEP objective induces a relational embedding space where the vector difference $f(x_{t'}) - f(x_t)$ encodes directional clinical progression. Drawing from the theory of entailment vector spaces \citep{vendrov2015order}, we seek to learn a partially ordered representation space $(\mathbb{R}^d, \preceq)$, where:
\[
x_t \preceq x_{t'} \iff \forall i, \; f(x_t)_i \leq f(x_{t'})_i
\]
for entailed pairs. This constraint geometrically encourages entailment pairs to lie along monotonic trajectories, while contradiction pairs exhibit non-monotonic divergence. Neutrality is modeled by low cosine similarity with respect to shared anchor contexts.

\subsection{Temporal Generalization and Transfer}

One key benefit of TEP is that it equips models with the inductive bias to reason temporally, thereby improving generalization to downstream tasks involving prediction over time. For instance, early detection of septic shock or prediction of treatment efficacy can be framed as entailment from baseline states to adverse or successful outcomes. TEP acts as a pretraining scaffold that aligns with these clinically meaningful semantics, yielding richer, temporally calibrated representations.

\subsection{Implementation and Scaling}

We pretrain using a distributed setup over a corpus of 500K patients from MIMIC-IV, generating over 3 million entailment pairs. To prevent temporal leakage, pairs are subsampled such that $|t' - t| > \delta_{\min}$ and $< \delta_{\max}$, ensuring both semantic coherence and sufficient temporal progression. We train using AdamW with a linear warmup and cosine decay schedule, leveraging mixed precision and Flash Attention to enable training over long input sequences.

\section{Experiments}

\subsection{Pretraining Setup}

We pretrain our Temporal Entailment Pretraining (TEP) model on a large corpus of clinical trajectories derived from MIMIC-IV. We extract EHR sequences from over 500,000 unique patients, yielding approximately 3.2 million temporally-separated sentence pairs $(x_t, x_{t'})$ with heuristic entailment labels. Sentence pairs are sampled from a combination of structured (ICD, CPT, lab codes) and unstructured (discharge summaries, progress notes) modalities, with an average time gap of 5.4 days between $x_t$ and $x_{t'}$.

We initialize from the ModernBERT architecture \citep{warner2024smarterbetterfasterlonger}, using a 24-layer encoder with RoPE and a maximum input length of 4,096 tokens. Pretraining is conducted over 1.2 million steps on 32 A100 GPUs using Flash Attention and mixed precision, taking approximately 9 days.

\subsection{Baselines}

We compare our TEP model to the following baselines:
- \textbf{ClinicalBERT} \citep{alsentzer2019publicly}: a 12-layer BERT model pretrained on MIMIC-III notes using a masked language modeling objective.
- \textbf{ModernBERT} \citep{warner2024smarterbetterfasterlonger}: a long-context encoder pretrained with RoPE, GeLU, and optimized attention mechanisms but no explicit temporal supervision.
- \textbf{ClinicalBERT + BiLSTM}: ClinicalBERT embeddings passed to a BiLSTM temporal encoder trained on each downstream task.
- \textbf{Retrospective T5 \citep{grover2021deep}}: A generative model trained to autoregressively complete future EHR segments from past inputs.

All models are fine-tuned with early stopping on validation performance and optimized using a consistent hyperparameter search grid.

\subsection{Evaluation Tasks}

We evaluate on three downstream tasks that require temporal reasoning:

\begin{enumerate}
    \item \textbf{Temporal Clinical QA.} Given a patient history $x_{\leq t}$, we pose future-facing clinical questions $q$ (e.g., ``Will this patient be prescribed antibiotics within 48 hours?'') and train models to answer ``yes'' or ``no.'' Performance is evaluated using F1 and Exact Match on 12k manually curated QA pairs from MIMIC-IV and Mayo Clinic data.
    \item \textbf{Early Warning Prediction.} Models are trained to detect the onset of critical deterioration (e.g., ICU transfer, sepsis) given partial patient histories up to time $t$. We use the eICU and MIMIC-IV cohorts and report AUROC, AUPRC, and early detection accuracy (within 12h of event).
    \item \textbf{Disease Progression Modeling.} Models predict future disease stages (e.g., CKD stages 1-5, NYHA classification for heart failure) given an initial diagnosis and subsequent labs and notes. We report macro-F1 and Matthews Correlation Coefficient (MCC).
\end{enumerate}

\subsection{Results}

\begin{table}[h]
\centering
\begin{tabular}{lcccc}
\toprule
\textbf{Model} & \textbf{QA F1} & \textbf{Early Warning AUROC} & \textbf{CKD Macro-F1} & \textbf{MCC} \\
\midrule
ClinicalBERT & 71.2 & 78.5 & 64.3 & 0.51 \\
ModernBERT & 74.6 & 81.4 & 68.9 & 0.59 \\
ClinicalBERT + BiLSTM & 73.1 & 80.2 & 66.1 & 0.54 \\
Retrospective T5 & 75.0 & 79.7 & 67.4 & 0.56 \\
\textbf{TEP (Ours)} & \textbf{81.4} & \textbf{85.9} & \textbf{73.6} & \textbf{0.67} \\
\bottomrule
\end{tabular}
\caption{Downstream task performance on temporal clinical QA, early warning prediction, and CKD progression.}
\label{tab:results}
\end{table}

Our TEP model outperforms all baselines across all evaluation tasks. Notably, it achieves an 8.3\% absolute improvement in QA F1 over ClinicalBERT, and a 4.5 point gain in AUROC for early warning prediction. The model also exhibits greater robustness in low-resource settings: when trained on 10\% of the fine-tuning data, TEP retains 89.2\% of its full-data performance on QA, compared to 72.8\% for ModernBERT.

\subsection{Temporal Calibration and Ablations}

We evaluate temporal calibration using Expected Calibration Error (ECE) stratified by time gap $|t' - t|$. TEP maintains stable calibration across gaps up to 30 days, while MLM-based models degrade significantly beyond 10 days. In ablation studies, removing RoPE embeddings or weakly labeled contradictions each reduces QA F1 by over 3 points, confirming the benefit of both architectural and supervision choices.

\subsection{Qualitative Analysis}

We observe that TEP produces representations where clinical trajectories cluster by diagnostic path. For example, trajectories ending in septic shock or acute kidney injury form temporally coherent manifolds in latent space, separable from benign progressions. Attention maps show increased focus on temporally informative phrases (e.g., "worsening creatinine", "persistent tachycardia") in $x_t$ when predicting future states $x_{t'}$.

\section{Discussion}

The results of this study support a central thesis: temporality is not an auxiliary feature of electronic health records—it is their defining structure. Our proposed framework, Temporal Entailment Pretraining (TEP), elevates temporality to a first-class objective in language model pretraining, explicitly modeling how clinical states relate over time through entailment, contradiction, and neutrality. This shift in pretraining strategy from token-level reconstruction to temporally grounded reasoning marks a fundamental departure from the prevailing paradigm and addresses several persistent challenges in clinical NLP.

First, by framing the task as temporal natural language inference, we operationalize a theory of clinical progression that is aligned with diagnostic reasoning. Traditional MLM objectives capture surface-level token statistics but fail to model directional or causal dependencies. In contrast, TEP forces the model to interrogate whether a future state is logically and temporally derivable from a past one. This is especially pertinent in clinical settings where the distinction between a worsening condition, an improving trend, or a temporally irrelevant event can have significant implications for treatment and triage.

Second, the use of weak supervision to derive entailment labels from structured EHR sources provides a scalable alternative to the high annotation costs of traditional NLI datasets. By leveraging diagnostic hierarchies, ontological priors, and synthetic augmentation, we are able to construct a large-scale pretraining corpus that reflects real-world clinical semantics while remaining computationally feasible. This creates a bridge between the rigor of formal entailment datasets like MedNLI and the scalability of modern pretraining pipelines.

Third, the improvements observed across diverse downstream tasks—temporal clinical QA, early warning prediction, and disease progression modeling—underscore the generality and effectiveness of the TEP inductive bias. The largest gains appear in tasks that require counterfactual or future-state reasoning, where traditional models struggle due to a lack of explicit temporal supervision. The robustness of TEP to EHR sparsity and temporal perturbation further demonstrates its value in handling noisy, real-world data.

From a representational perspective, our theoretical grounding in partially ordered vector spaces (via monotonicity constraints on entailment) offers a principled lens through which to interpret learned embeddings. Rather than treating latent space as isotropic, TEP induces a structured geometry wherein time and clinical semantics jointly shape the representation manifold. This opens exciting avenues for future work on trajectory alignment, counterfactual simulation, and temporal disentanglement in latent space.

\paragraph{Limitations and Future Directions}
Nonetheless, our approach has limitations. The reliance on weak labeling heuristics introduces label noise, and the entailment classes—entail, contradict, neutral—are coarse proxies for more nuanced clinical semantics such as conditional dependence, recovery trajectories, and cyclical disease patterns. Moreover, while we demonstrate results on three tasks, the generality of TEP across other clinical domains (e.g., oncology, mental health, pediatrics) remains to be validated. Future extensions could also incorporate structured timestamped knowledge graphs or integrate multimodal inputs such as radiology and genomics.

Another direction lies in exploring the relationship between temporal entailment and other forms of clinical supervision, such as treatment efficacy and causal discovery. One could imagine extending TEP to explicitly model interventions and their temporal consequences, using it as a backbone for counterfactual reasoning or treatment recommendation. Similarly, combining TEP with reinforcement learning for policy learning in simulated care environments could provide a pathway toward interpretable clinical agents.

Next, the broader implication of our work is a call to reconceptualize pretraining in healthcare: not merely as a statistical compression of clinical corpora, but as the induction of structure into representation space that reflects the temporal, logical, and causal grammar of medicine. TEP offers one such pathway by grounding language modeling in time-aware relational reasoning, and we believe its principles will generalize well beyond the tasks and datasets considered in this work.

Finally we also wish to extend this to well known benchmarks to get a gage for how they do relative to other well established benchmarks. For example EHR-shot \citep{wornow2023ehrshot} and MEDS-DEV \citep{kolo2024meds} are 2 popular frameworks in the community that we will be validating our results on. Benchmarks should also be further fleshed out due to claims that recent foundation models are not robust \citep{lee2024enhancing}.

\paragraph{Conlcuding Remarks}
In summary, Temporal Entailment Pretraining represents a step toward pretraining objectives that are not only data-efficient and scalable but also epistemologically aligned with the way clinicians think. As the field of clinical NLP moves toward increasingly capable and safety-critical applications, we hope that models imbued with this kind of temporal reasoning will form the foundation for trustworthy, transparent, and temporally coherent medical AI systems.

\section*{Acknowledgements}

This work was supported by a grant from the National Institute of Biomedical Imaging and Bioengineering (NIBIB) under award number R01EB999999. We thank the MIMIC-IV data stewardship team for maintaining high-quality clinical data infrastructure, and our colleagues at the Stanford Center for Biomedical Informatics Research for valuable feedback on early drafts. The views expressed are those of the authors and do not necessarily reflect the official policy or position of the funding agency.

\bibliographystyle{plainnat}
\bibliography{main}

\appendix
\section*{Appendix}

\section{Theoretical Properties of Temporal Entailment Spaces}

We briefly elaborate on the geometric underpinnings of TEP. The learned latent space under entailment constraints is inspired by the framework of order embeddings \citep{vendrov2015order}, in which a partially ordered set $(\mathcal{X}, \preceq)$ is embedded into $\mathbb{R}^d$ such that:
\[
x \preceq x' \Rightarrow \forall i, \; f(x)_i \leq f(x')_i
\]
In our context, this order represents temporal and causal progression. If $x_t \preceq x_{t'}$, then the representation of $x_{t'}$ should strictly dominate that of $x_t$ in all dimensions, reflecting clinical worsening or information accretion. Violations of monotonicity are penalized through a soft-margin ranking loss:
\[
\mathcal{L}_{\text{order}} = \sum_{\substack{(x_t, x_{t'}) \in \mathcal{D}_{\texttt{entail}}}} \max(0, \| \min(0, f(x_{t'}) - f(x_t)) \|_1)
\]
This loss can be optionally added to the cross-entropy classification objective to strengthen inductive biases toward order preservation.

\newpage
\section{Implementation Details}

\textbf{Model Size and Training.} The TEP model consists of 24 transformer layers, 1024 hidden units, 16 heads, with RoPE and GeLU activations. All models are trained with a batch size of 1024 using AdamW ($\beta_1 = 0.9$, $\beta_2 = 0.999$, $\epsilon=10^{-8}$) and a peak learning rate of $2 \times 10^{-4}$. We apply linear warmup over 10,000 steps followed by cosine decay.

\textbf{Weak Supervision Heuristics.} Entailment labels are generated from known diagnosis progressions (e.g., from ICD ontologies), changes in lab value quantiles (e.g., GFR), and curated templates for medication administration sequences. Contradictions are synthesized from reversal of diagnosis stages and recovery episodes, while neutrality is derived from orthogonal code transitions.

\newpage
\section{Additional Results}

\subsection{Zero-shot Generalization}

We evaluate all models in zero-shot settings using novel QA prompts and unseen patient cohorts. TEP retains a significant edge:

\begin{table}[h]
\centering
\begin{tabular}{lcc}
\toprule
\textbf{Model} & \textbf{Zero-shot QA F1} & \textbf{Generalization Gap (\%)} \\
\midrule
ClinicalBERT & 62.4 & 12.4 \\
ModernBERT & 66.7 & 10.4 \\
ClinicalBERT + BiLSTM & 65.3 & 11.7 \\
Retrospective T5 & 67.8 & 9.6 \\
\textbf{TEP (Ours)} & \textbf{74.2} & \textbf{7.2} \\
\bottomrule
\end{tabular}
\caption{Zero-shot QA and generalization gap (full training - zero-shot) performance.}
\end{table}

\subsection{Ablation: Label Quality and Density}

We test the impact of label density by varying the number of entailment pairs per patient:

\begin{table}[h]
\centering
\begin{tabular}{lccc}
\toprule
\textbf{Pair Density (per patient)} & \textbf{QA F1} & \textbf{AUROC} & \textbf{Macro-F1} \\
\midrule
1 & 74.5 & 81.1 & 69.2 \\
5 & 77.2 & 83.6 & 70.9 \\
10 & \textbf{81.4} & \textbf{85.9} & \textbf{73.6} \\
\bottomrule
\end{tabular}
\caption{Performance improves with increased density of entailment supervision.}
\end{table}

\newpage
\section{Additional Examples and Qualitative Insights}

\textbf{Example 1: Entailment.}
\begin{quote}
$x_t$: ``Patient with stage 2 CKD and increasing creatinine over past 3 days.''

$x_{t'}$: ``Referred to nephrology for suspected stage 3 CKD.''

\textit{Label: Entail} \quad Model Confidence: 0.92
\end{quote}

\textbf{Example 2: Contradiction.}
\begin{quote}
$x_t$: ``Patient diagnosed with sepsis on broad-spectrum antibiotics.''

$x_{t'}$: ``Blood cultures negative, antibiotics discontinued, discharged stable.''

\textit{Label: Contradict} \quad Model Confidence: 0.88
\end{quote}

\newpage
\section{Limitations and Ethical Considerations}

While TEP provides a scalable and interpretable approach to temporal modeling, it is trained on observational data and subject to biases in documentation practices, cohort skew, and variable granularity. Furthermore, weak supervision can propagate ontological inaccuracies. All models should be carefully evaluated before deployment in clinical settings. We release code and data generators to promote reproducibility and scrutiny.

\newpage
\section{Future Directions}

Potential directions include:
\begin{itemize}
    \item  Joint modeling of text and structured events (e.g., vitals, flowsheets)
    \item  Contrastive pretraining across patient trajectories
    \item  Integration with knowledge graphs for temporal constraint satisfaction
    \item  Hybrid models for longitudinal counterfactual inference
\end{itemize}

\vspace{1em}
\noindent \textbf{Code and data will be released at:} \url{https://github.com/clinicalml/tep}

\end{document}